\documentclass[]{llncs}
\usepackage{graphicx}
\usepackage{amsmath,graphicx}
\usepackage{tabularx}
\usepackage{listings}
\usepackage{multirow}
\usepackage{makecell}
\usepackage{geometry}
\begin{document}
\title{MesoGraph: Automatic Profiling of Malignant Mesothelioma Subtypes from Histological Images}
%
\author{Mark Eastwood \inst{1} \and 
Heba Sailem\inst{2,3} \and
Silviu Tudor Marc\inst{4} \and 
Xiaohong Gao\inst{4} \and
Judith Offman\inst{5} \and
Emmanouil Karteris\inst{6} \and
Angeles Montero Fernandez\inst{7} \and
Danny Jonigk\inst{8} \and
William Cookson\inst{9} \and
Miriam Moffatt\inst{9} \and
Sanjay Popat\inst{9} \and
Fayyaz Minhas\inst{1,10} \and
Jan Lukas Robertus\inst{9,10}}

\institute{Tissue Image Analytics Center, University of Warwick\and
Institute of Biomedical Engineering, University of Oxford\and
School of Cancer and Pharmaceutical Sciences, King's College London\and
Dept. of Computer Science, University of Middlesex\and
Queen Mary University of London\and
College of Health, Medicine and Life Sciences, Brunel University London\and
Manchester University\and
Medizinische Hochschule Hannover\and
National Heart and Lung Institute, Imperial College London\and
Joint last authors}
%
%
\maketitle              
%
\section*{Summary}
Malignant mesothelioma is classified into three histological subtypes, Epithelioid, Sarcomatoid, and Biphasic according to the relative proportions of epithelioid and sarcomatoid tumor cells present. Biphasic tumors display significant populations of both cell types. Current guidelines recommend that the sarcomatoid component of each mesothelioma is quantified, specifically, as a higher percentage of sarcomatoid pattern in biphasic mesothelioma shows poorer prognosis. It is therefore important to accurately diagnose morphological subtypes. However, this subtyping is subjective and limited by current diagnostic guidelines. This can affect the reproducibility even between expert thoracic pathologists when characterising the continuum of relative proportions of epithelioid and sarcomatoid components using a three class system. In this work, we develop a novel dual-task Graph Neural Network (GNN) architecture with ranking loss to learn a model capable of scoring regions of tissue down to cellular resolution. This allows quantitative profiling of a tumor sample according to the aggregate sarcomatoid association score of all the cells in the sample. The proposed approach uses only core-level labels and frames the prediction task as a dual multiple instance learning (MIL) problem. Tissue is represented by a cell graph with both cell-level morphological and regional features. We introduce a dataset of 234 mesothelioma tissue micro-array cores, and use an external multi-centric test set from Mesobank, on which we demonstrate the predictive performance of our model. We validate our model predictions through an analysis of the typical morphological features of cells according to their predicted score, finding that some of the morphological differences identified by our model match known differences used by pathologists. We further show that the model score is predictive of patient survival with a hazard ratio of 2.30. The ability to quantitatively score tissue on how much sarcomatoid component is present, and overlay cell-level instance scores, will assist pathologists in assessing a sample more quickly and consistently by defining the sarcomatoid component, particularly in biphasic mesothelioma. This ultimately will improve treatment decisions for patients with mesothelioma. The code for the proposed approach, along with the dataset, is available at: https://github.com/measty/MesoGraph.

Our contributions include releasing a new dataset, introduction of the MesoGraph architecture providing accurate  subtype predictions, and insightful analysis of morphological differences between sarcomatoid and epithelioid components.
\keywords{Graph Neural Networks  \and Multiple Instance Learning \and Mesothelioma \and Cancer subtyping \and Digital Pathology}
%
%


\section{Introduction} \label{intro}

Malignant mesothelioma (MM) is an aggressive cancer of malignant mesothelial cells of the pleural lining, primarily associated with asbestos exposure \cite{WAGNER1960a}. It has a poor prognosis with less than 10\% five year survival rates due to late diagnosis (\cite{hjerpe2018cyto} ; \cite{mesostats}). It has a long latency period from initial exposure to eventual carcinogenesis, and is difficult to diagnose due to its nonspecific clinical manifestations. MM is classified into 3 subtypes \cite{Ai2014}, epithelioid, biphasic and sarcomatoid mesothelioma (EM, BM and SM respectively), with BM characterised by a mix of epithelioid and sarcomatoid components. 
The Histological subtype of malignant mesothelioma is essential for prognosis and clincial decisions on treatment pathways for patients \cite{Meyerhoff2015ImpactOM}. Stratification of a given sample into a particular subtype informs treatment and can help gain a more in-depth understanding of disease pathology and outcome. The benefit of surgical treatment has prognostic implications for epithelioid mesothelioma with median survival of 19 months but less so for SM and BM with respective median survival of 4 months and 12 months after surgical treatment \cite{Mansfield2014SystematicRO}. 
 
Epithelioid mesothelioma is characterised by malignant cells that are cytologically round with varying grading of atypia. Sarcomatoid mesothelioma cells are generally recognised as malignant elongated spindle cells \cite{whotumors} and are associated with worse prognosis in comparison to EM. Sarcomatoid mesothelial cells may also include transitional features that are intermediate between epithelioid and sarcomatoid. Although transitional cells are now classified under sarcomatoid mesothelioma, their presence is associated with worse prognosis \cite{courtiol2019meso}.
 
While the distinction of these three histological subtypes of MM is crucial to patient treatment, management and prognosis, it is challenging to differentiate  EM, SM and BM through visual analysis. Currently, there are no clear guidelines on how to perform this stratification in an objective and reproducible manner \cite{dacic2020interobserver}. Furthermore, even though mesotheliomas are divided into these three broad categories, in reality there is a continuous spectrum from EM to SM dependent upon the relative proportion of epithelioid and sarcomatoid cells in a given sample. As a consequence, existing approaches are unable to objectively quantify where on this spectrum a given sample falls based on profiling of cellular morphological patterns in it. 

A number of deep learning methods for analyzing mesothelioma images have been developed recently. For example, SpindleMesoNET~\cite{Naso2021} can separate malignant SM from benign spindle cell mesothelial proliferations. A recent approach for survival prediction of MM patients called MesoNet~\cite{courtiol2019meso} uses a Mulitple Instance Learning (MIL) solver originally developed for computer vision applications~\cite{courtiol2018weak} and classification of lymph node metastases \cite{weldon2016}. However, automated subtyping of malignant mesothelioma from hematoxylin and eosin (H\&E) stained tissue sections remains an open problem.

\begin{figure*}[ht]

  \centering
 \centerline{\includegraphics[width=15.5cm]{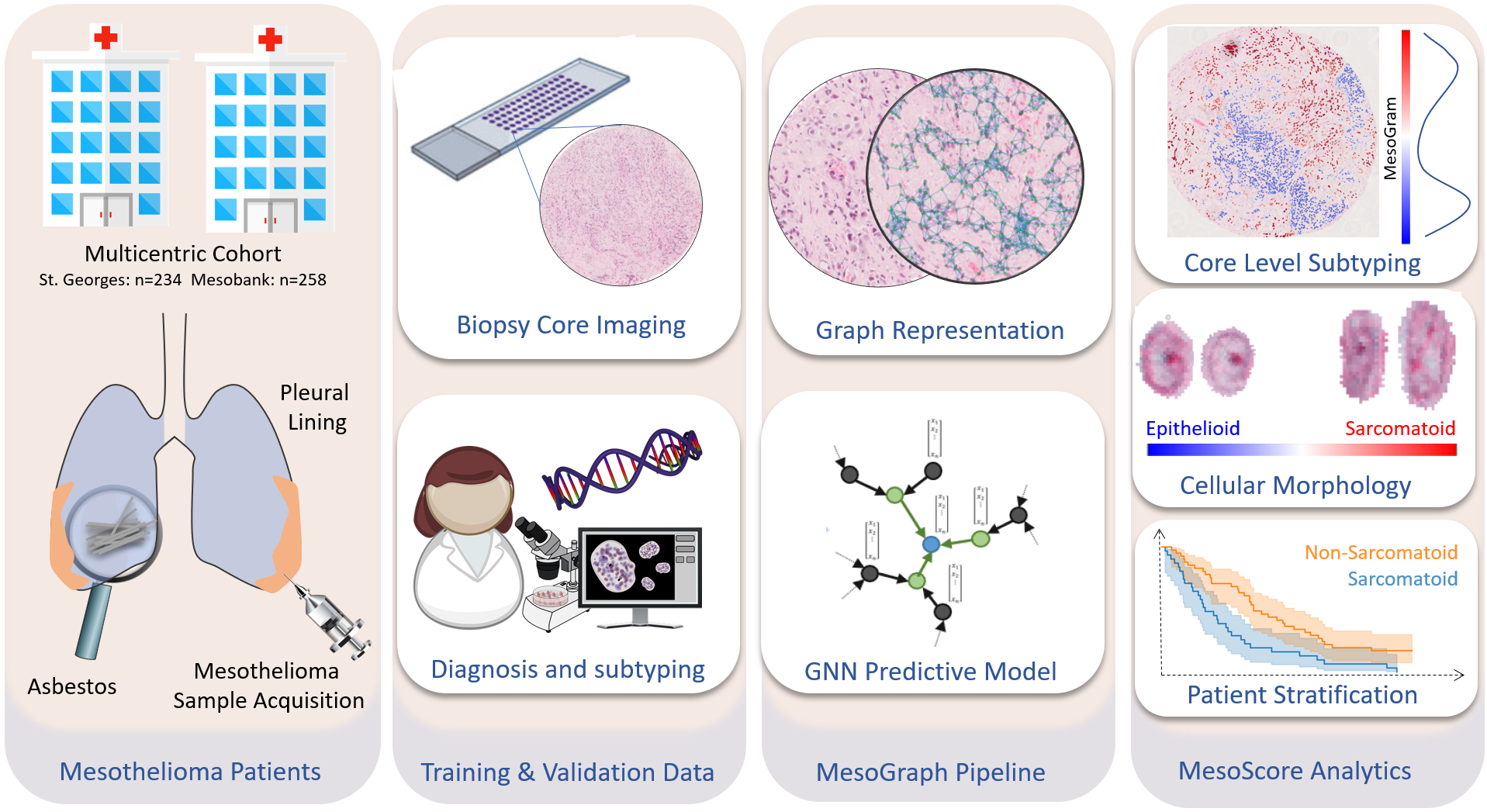}}

\caption{High level overview/graphical abstract of our approach to mesothelioma subtyping. Pathologist labelled TMA cores are used to learn a graph nueral network model on cell-level features. We show the model can identify sarcomatioid areas, and its predicted scores can be used to stratify patient survival.}
\label{graphical_abs}
\end{figure*}

One challenge in the characterization of mesothelioma is that pathologist-assigned ground-truth labels of mesothelioma subtypes are typically available only at the case level as it is very difficult for pathologists to associate tumor micro-environment  or cellular morphometric patterns with image or case level labels in an objective manner. Moreover, it can be very time-consuming to obtain detailed cellular or regional annotations, and those annotations may not be very reliable due to significant inter- and intra-observer variation. 

The aim of this study was to develop a GNN approach to predict subtypes of malignant mesothelioma in a MIL setting. This was achieved considering Tissue Micro-array (TMA) cores as bags and individual cells as instances.
On these we have built a weakly-supervised machine learning model to characterize mesothelioma subtypes using only case level labels in its training. The proposed approach can generate a quantitative assessment of where the sample stands in terms of the aforementioned epithelioid to sarcomatoid spectrum, enabling pathologists to perform a more in-depth characterization of tumor samples. A high level overview of the approach presented in the paper can be found in Fig. \ref{graphical_abs}.


\section{Results and Discussion}
We have developed a custom Graph Neural Network based pipeline called MesoGraph that can predict mesothelioma subtypes using H\&E stained tissue images. MesoGraph uses pathologist assigned case-level labels without any cellular or regional annotations in its training. The proposed approach models each cell in a given sample as a node in the graph, which is connected to neighbouring cells. Each node is associated with various features, which can be broadly classified into four types: 1) nuclear morphometric features, 2) stain intensity features of nuclear and cytoplasmic components of the cell, 3) cellular counts in the neighbourhood of node and 4) deep neural network and haralick based texture features. For a given test sample, it generates two probability scores (collectively called MesoScore) representing the probabilities of the sample being epithelioid or sarcomatoid. As a biphasic mesothelioma tumor is composed of both epithelioid and sarcomatoid components, the two outputs in MesoScore allow precise quantification of the two components in the sample. In addition to predicting mesothelioma subtype, MesoGraph also generates cell-level quantitative scores representing the association of each cell with the mesothelial subtype of the given sample. MesoGraph has been trained and independently validated on two datasets: St. George's Hospital (SGH, n = 234) and the multi-centric MesoBank collection (MB, n = 258). 

In this section, we present the results of the proposed method in terms of its predictive performance in comparison to existing approaches, as well as its ability to identify histological features and morphological characteristics of cells associated with different types of malignant mesothelioma. We also demonstrate the ability of the proposed approach to stratify mesothelioma patients based on their expected survival. 

\subsection{Predictive Performance} \label{performance_results}
\begin{figure*}[ht]

  \centering
 \centerline{\includegraphics[width=14cm]{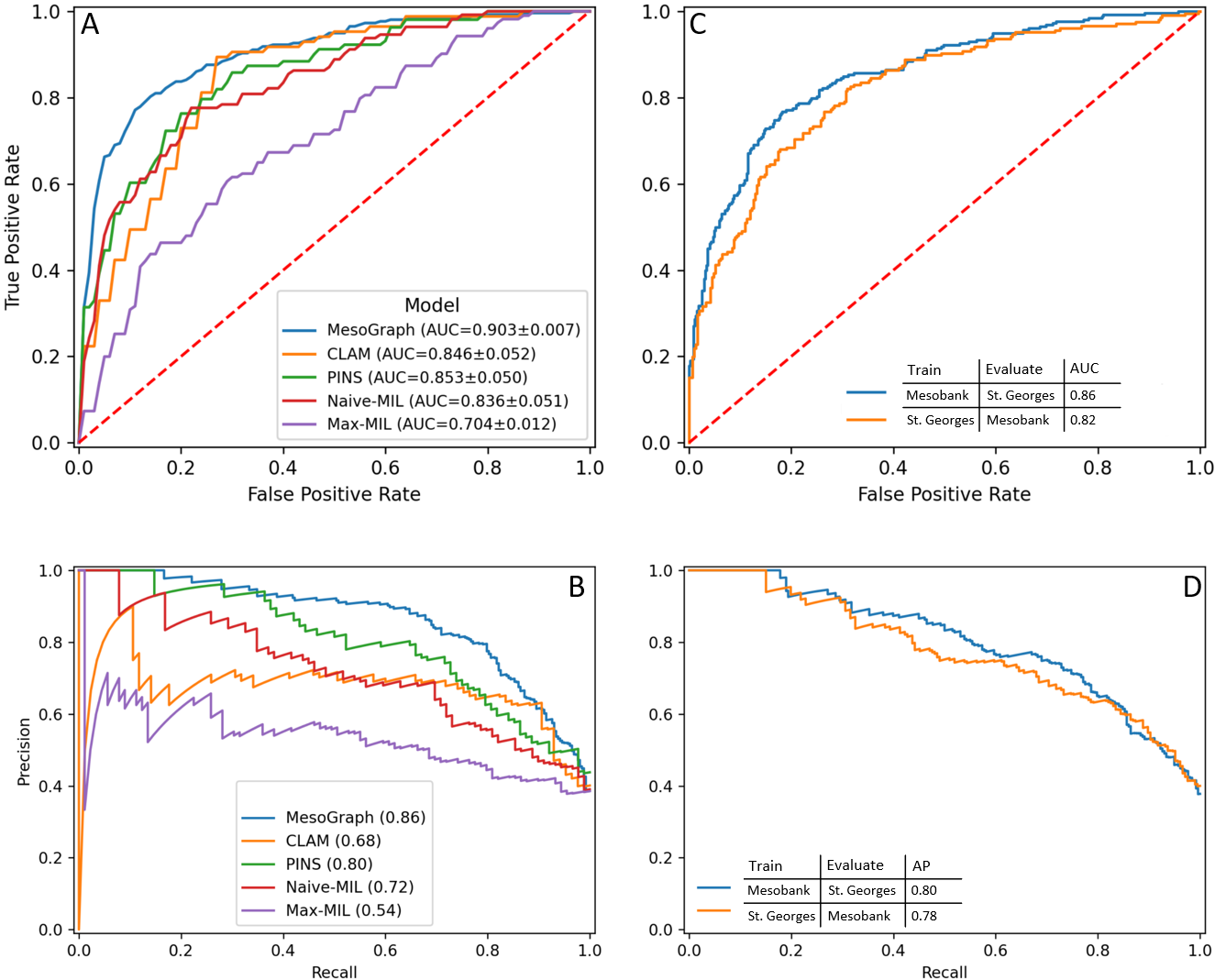}}

\caption{A) AUROC for each model. B) Model average precision. C) AUROC on cross-cohort evaluation. D) Average precision on cross-cohort evaluation. Our model outperforms other models tested, and from C and D we see good generalization to external validation data.}
\label{model_auc}
\end{figure*}

Test results from the MesoGraph pipeline for both MesoBank and SGH datasets are shown in Figure \ref{model_auc} and Table~\ref{restable}. Here, the Receiver Operating Characteristic (ROC) curve is obtained by considering both sarcomatoid and biphasic samples as the positive class, whereas the epithelioid samples are associated with the negative label. As can be seen from these results, the proposed approach offers high predictive quality over both cross-validation and independent testing in comparison to other existing approaches. In the table, PINS refers to the Positive Instance Sampling Patch-based MIL approach as detailed in \cite{eastwood2022pins} whereas CLAM is the Clustering-constrained Attention Multiple Instance Learning method described in \cite{lu2020data}, a deep-learning-based weakly-supervised method that uses attention in combination with clustering-based constraints to identify the most predictive areas of the image.  Max-MIL and naive-MIL are simple patch based baseline multiple instance learning (MIL) methods detailed further in the methods Section \ref{evaluation}.
As can be seen from Table \ref{restable}, the max-based MIL strategy performs poorly. This is likely due to the relatively small size of the training dataset, as learning only on the maximally scoring instance per bag exacerbates this.
Naive MIL performs surprisingly well. This may be due to the relatively high proportion of positive instances that are expected to be present in many of the positive bags (for example a sarcomatoid core should contain mostly positive instances).
This makes the implicit assumption this model makes, namely that all instances share the label of the bag, less wrong for this dataset compared to other MIL tasks.
PINS and CLAM, as two patch based methods with a mechanism for focussing on the most relevant region of an image, perform similarly with solid performance. However, as patch based methods, the spatial resolution of the prediction maps they can provide is lower than that of our cell graph based model.

Our model outperforms other models tested, achieving an internal cross-validation performance of $0.90 \pm 0.01$. It performs particularly well in terms of its average precision (AP) of $0.86\pm0.02$, indicating that its performance on the positive class (which is the minority class) is very good. While performance drops slightly on the external validation set, an AUROC of $0.86$ and AP of $0.8$ as seen in Fig \ref{model_auc}, panels C and D shows these results generalize well. We attribute the performance improvements achieved by our model firstly to the cell graph representation, with cells and their morhpological features as instances, which is far more natural than an arbitrary division into patch instances. Secondly, our formulation of the learning as a dual task problem with a ranking loss. This acknowledges both the ordering we know exists between EM, BM and SM cores in terms of how much sarcomatoid component is present, and also the possibility that some regions of tissue may not be strongly associated with either a sarcomatoid or epithelioid core label.

\begin{table*}[t]
\begin{center}
\begin{tabularx}{0.85 \textwidth}{|>{\raggedright\arraybackslash}X||>{\raggedright\arraybackslash}X|>{\raggedright\arraybackslash}X|>{\raggedright\arraybackslash}X|>{\raggedright\arraybackslash}X|}

\hline
Metric&AUC-ROC&Avg. Precision&Sensitivity&Specificity\\
\hline\hline

max-MIL&$0.70\pm0.01$&$0.54\pm0.12$&$0.54\pm0.07$&$0.73\pm0.09$\\
naive-MIL&$0.84\pm0.05$&$0.72\pm0.11$&$0.72\pm0.08$&$0.71\pm0.1$\\
PINS \cite{eastwood2022pins}&$0.85\pm0.05$&$0.80\pm0.07$&$0.82\pm0.1$&$0.71\pm0.13$\\
CLAM \cite{lu2020data}&$0.85\pm0.07$&$0.74\pm0.11$&$0.75\pm0.11$&$0.77\pm0.02$\\
\small{MesoGraph(Ours)}&$0.90\pm 0.007$&$0.86\pm0.02 $&$ 0.88\pm0.015$&$0.72\pm0.01 $\\

\hline
\end{tabularx}
\caption{Summary of results (mean$\pm$stdev) of models evaluated on primary dataset (SGH). } \label{restable}
\end{center}
\end{table*}
\renewcommand{\arraystretch}{1}

\subsection{Visualization of Model Output}

\begin{figure*}[ht!]

  \centering
 \centerline{\includegraphics[width=14cm]{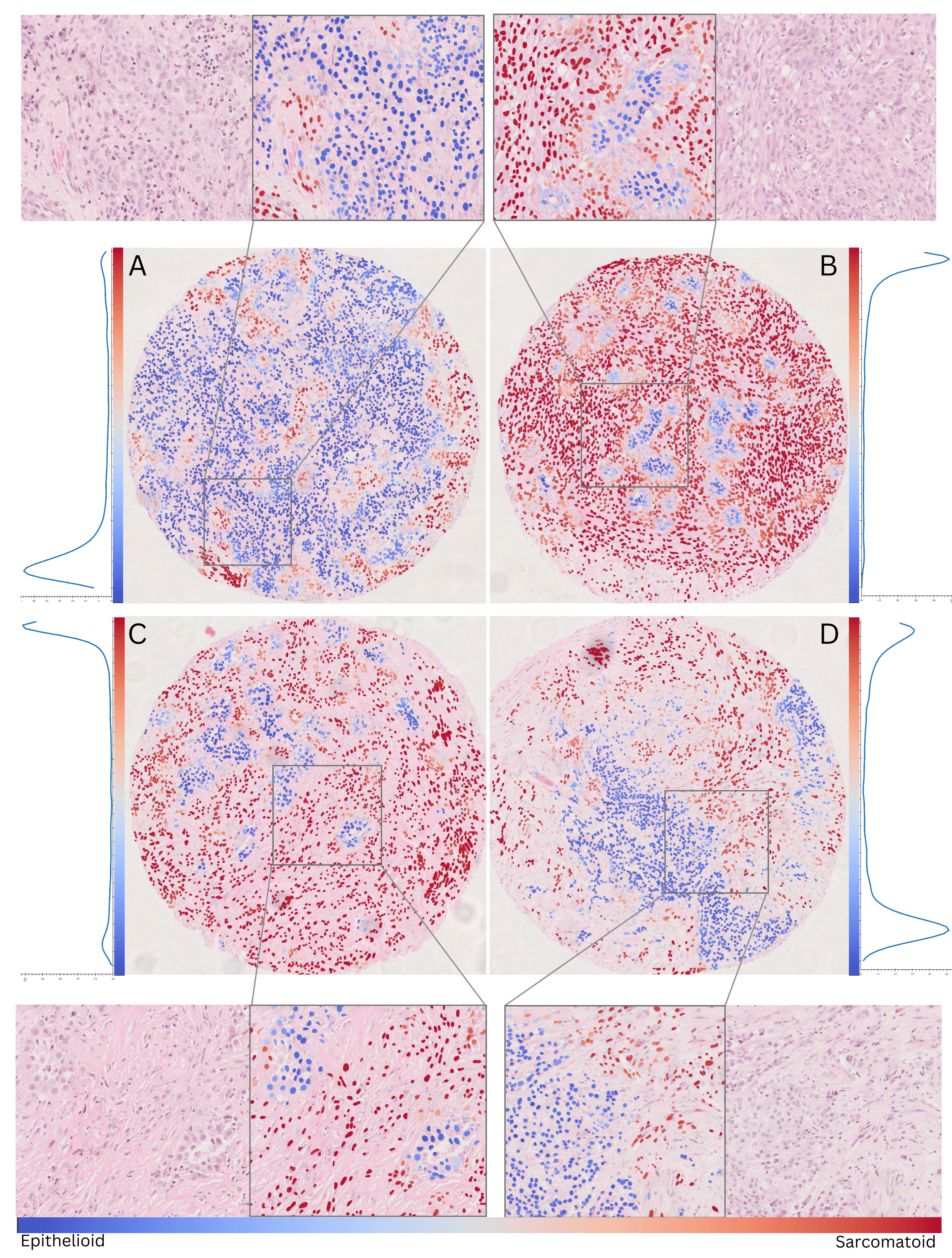}}

\caption{Model visualisation. Each core is shown together with a zoomed in view showing the differences in morphology between regions, and a plot showing the distribution of node scores in that core.  A. Epithelioid core. We see a predominantly low-scoring core with only a few small regions displaying slightly more sarcomatoid features. Conversely, in B, a Sarcomatoid core, nodes are predominantly high scoring. C and D. Biphasic cores. In each core we see a bimodal distribution of scores, particularly pronounced in core D. The zoomed in regions show a distinct difference in morphology between high and low scoring regions, with rounder cells seen in lower scoring regions, and a more elongated morphology and less structured cell organisation in higher scoring regions.}
\label{cores_vis}
\end{figure*}

The output of our model can be visualised in a zoomable, interactive Graphical User Interface (GUI) we have developed. A demo of results from our model can be found at (https://mesograph.dcs.warwick.ac.uk). Examples of the cell level scoring output by the model are shown in Fig. \ref{cores_vis}. For each cell in a given sample, the proposed model generates two prediction scores signifying the probability of the cell being associated with sarcomatoid or epithelioid labels. These scores can be combined and visualized in a colourmap showing cells that are associated with epithelioid (blue) and sarcomatoid (red) subtypes as well as a histogram (called MesoGram) showing the relative distributions of epithelioid and sarcomatoid components. 

\begin{figure*}[ht!]

  \centering
 \centerline{\includegraphics[width=14cm]{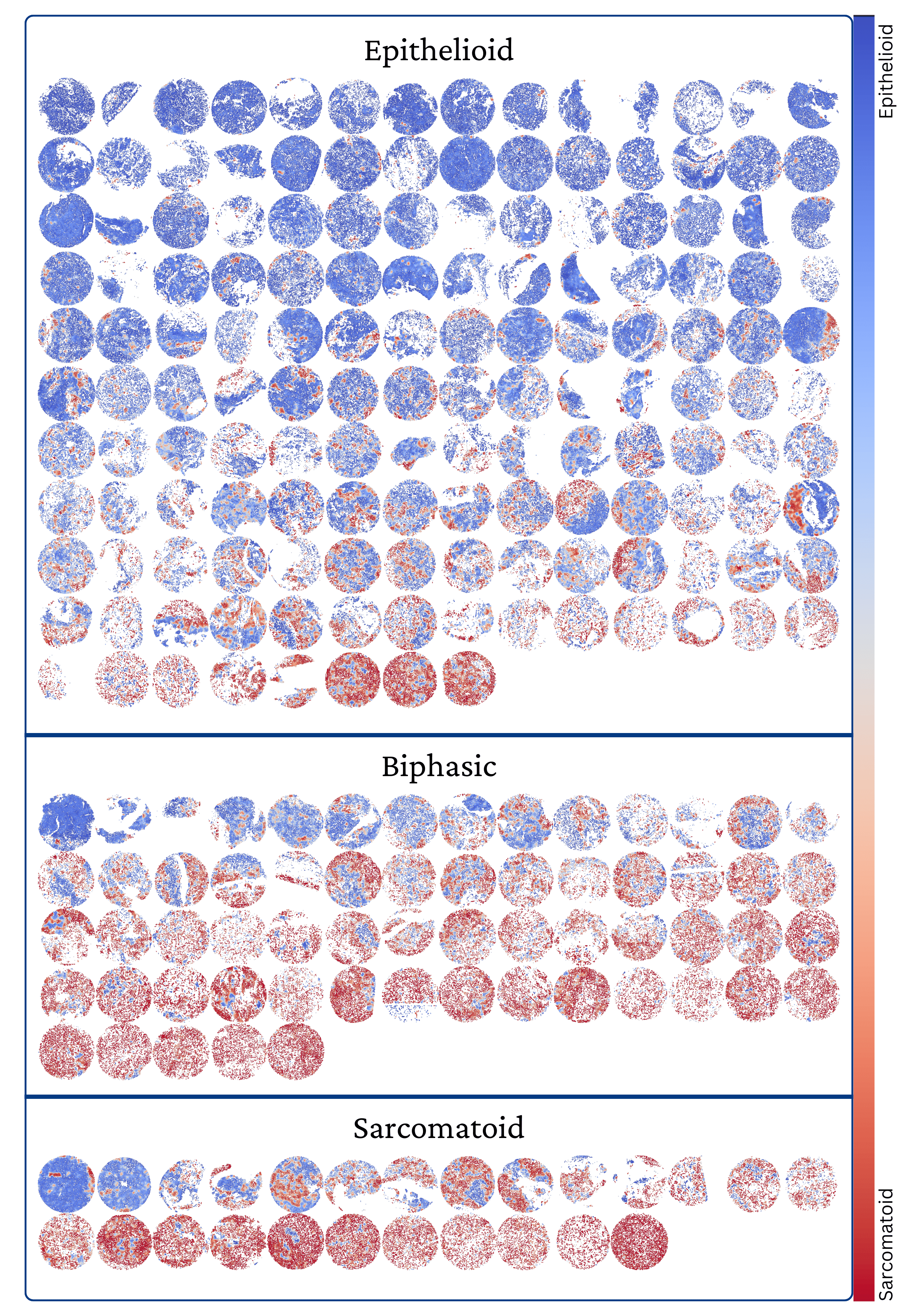}}

\caption{Overview of model predictions by subtype. Images of model predictions ordered within each subtype by the predicted predominance of the sarcomatoid component, illustrating the underlying continuous biological expression of tissue on the E to S spectrum.}
\label{cores_overview}
\end{figure*}

From the zoomed-in masks, we can see the model can distinguish between regions with typical rounded morphology of the Epithelioid subtype and the more elongated morphology displayed in Sarcomatoid regions. The MesoGram plots of most samples tend to be bimodal to some extent, with epithelioid and sarcomatoid cores more heavily skewed toward low and high scores, respectively. This continuum of distribution between sarcomatoid and epithelioid is demonstrated further in Fig. \ref{cores_overview}, where thumbnails of model output on all cores are shown, grouped by subtype and ordered within each subtype by model score. This ability to give a more precise, fine-grained characterisation of a tumor beyond the current three poorly defined and subjective subtypes is a strength of our approach.

\subsection{Explainability of model predictions}

\begin{figure*}[ht!]

  \centering
 \centerline{\includegraphics[width=14cm]{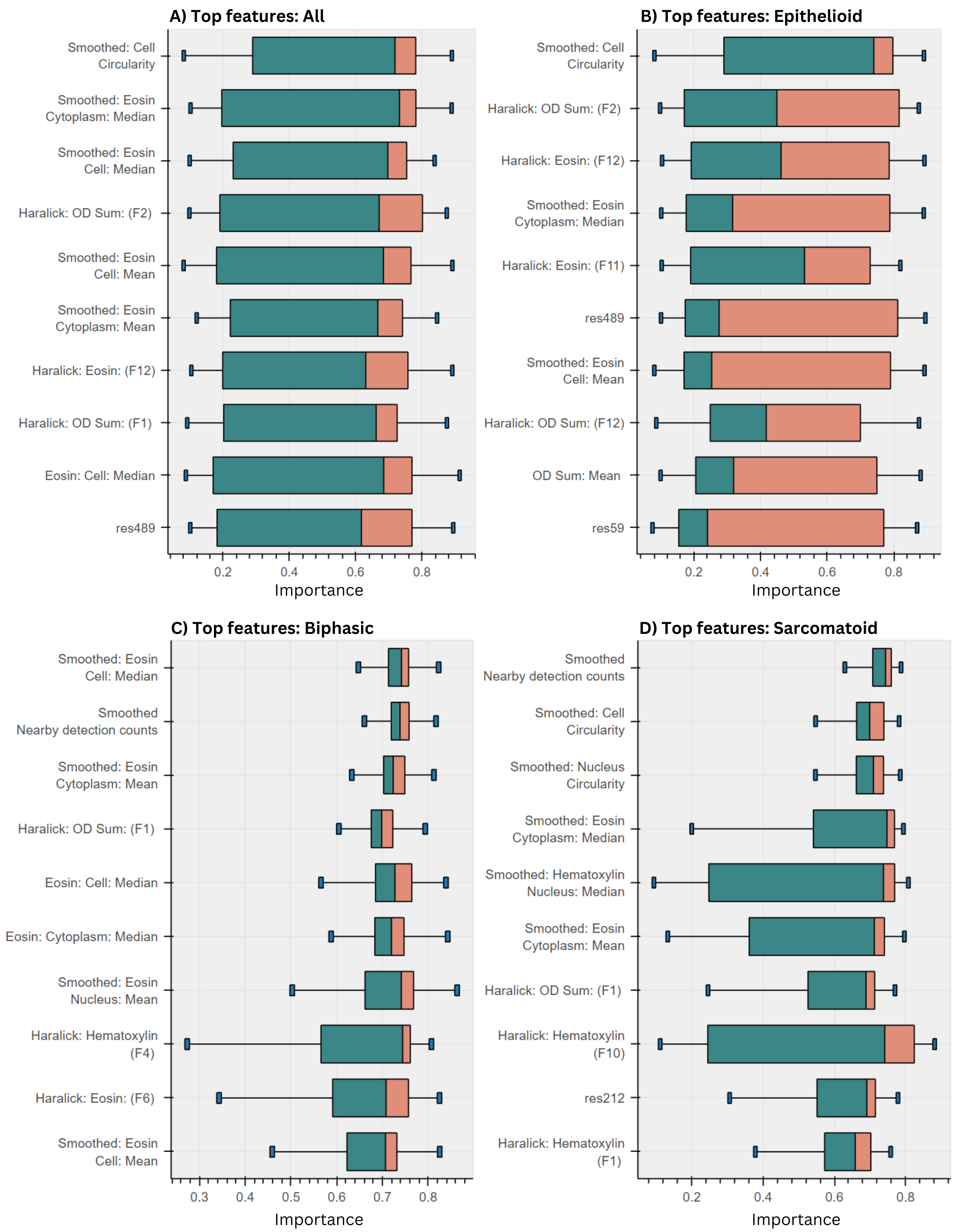}}

\caption{Top 10 features identified by GNNExplainer, considering A) all cores, and in B-D, importances on cores grouped by subtype. Results shown as a standard box \& whisker plot, with the box showing the 25th, 50th and 75th percentile of a features importance scores over cores. Whiskers show min and max values, limited at box $\pm$ 1.5 $\times$ inter-quartile range. The top feature is circularity, a knows differentiating characteristic between mesothelial subtypes, providing validation for our model.}
\label{feat_imp}
\end{figure*}

To gain an understanding of the predictions generated by the proposed approach, we have applied GNNExplainer \cite{ying2019gnnexplainer} to the trained graph neural network model. This allows us to understand what node level features are contributing to the prediction of a given sample for each subtype. The top 10 features identified in this analysis are shown in Fig. \ref{feat_imp}.

The most important feature overall is the circularity, which confirms the expected distinction between the rounder morphology of the epithelioid subtype, compared to the more spindle-shaped sarcomatoid morphology. There are also a number of features describing the intensity and texture in the Eosin channel around the nucleus. Looking at the feature importances on specific subtypes, the resnet features are most useful on epithelioid cores. Circularity is specifically important in Epithelioid and sarcomatoid subtypes, as they tend to be composed of more homogeneous populations and therefore are expected to contain mostly either rounded or more elongated cells. In both Biphasic and Sarcomatoid cores, nearby detection counts seem to be an important feature. This may reflect a tendency for non-epithelioid tumors to display a slightly more spread out and disorganised cell distribution. We also notice that the importance of most of the top features has far more spread when considering Epithelioid cores, indicating a wide variety of features can contribute to an epithelioid score, with few features being universally important across all epithelioid cores.

To determine the separation between classes based on top scoring features, in the lower half of Fig. \ref{morph_fig} (C and D) we plotted the prediction of each core against the assigned label by a pathologist. While Epithelioid cores and Sarcomatoid cores are mostly well-separated, we observe there can be overlap between some of the cases in terms of morphology. We also observe that Biphasic is not very distinct from Sarcomatoid cores.

\subsection{Characterization of Cellular Morphologies} \label{pca-umap}

\begin{figure}[ht!]  
\centerline{\includegraphics[width=15.8cm]{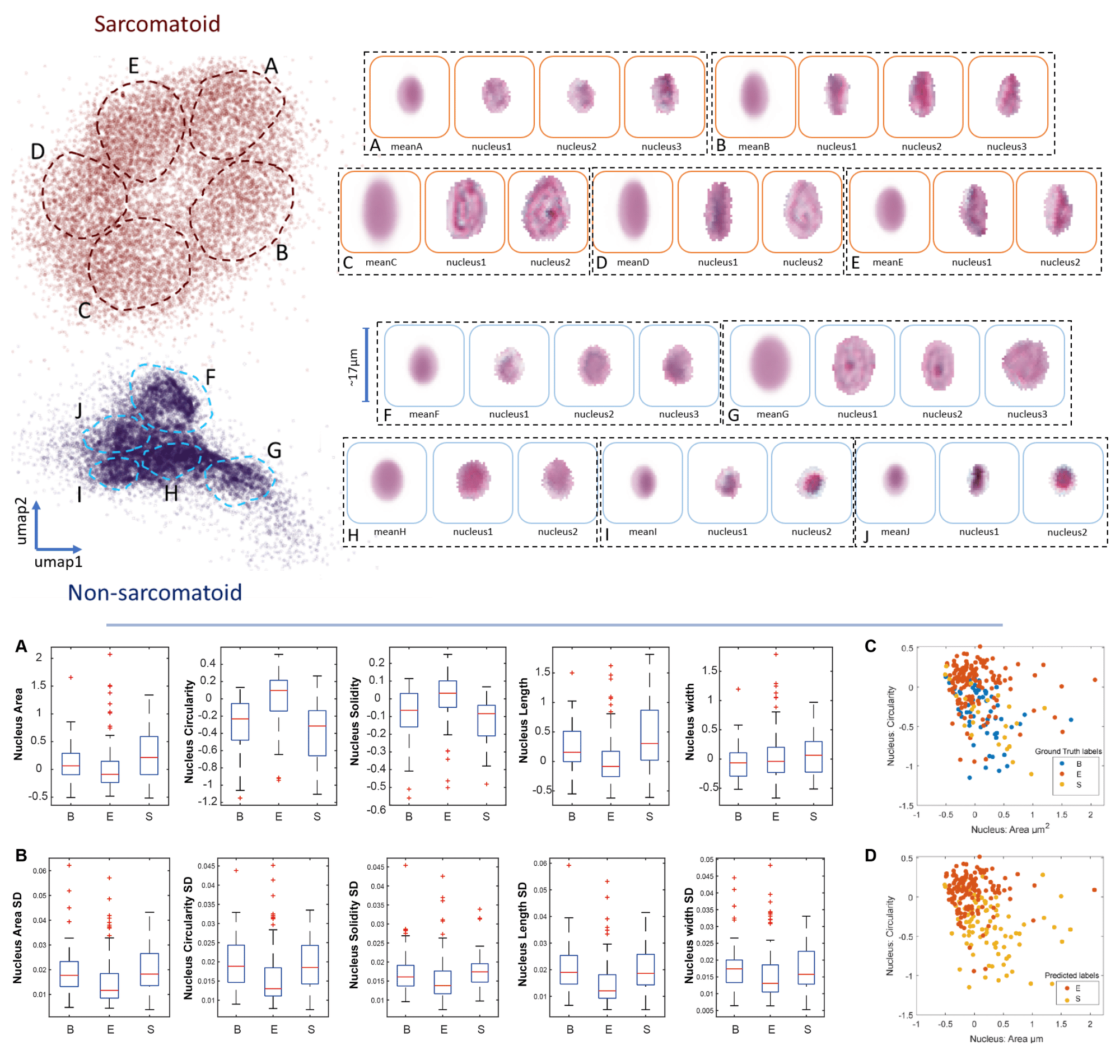}}
\caption{Upper panel: Examples of cells scored most and least highly by the model, plotted as a 2-D UMAP reduction
of principal components calculated on both high and low scoring cells. For each cluster, the mean of the cells is
displayed, together with individual example cells.
The clusters A-E, on the left side of the panel are predicted as Sarcomatoid and demonstrate a more spindle like morphology, grouped together in size and relative spindle cell characteristics in each cluster as shown by example cells on the right side of the panel. Similarly, the Non-sarcomatoid predicted cells also show clustering into 5 groups, F-J.
Lower panel: Morphological heterogeneity of mesothelioma tumours independent of model prediction.
A) Distribution of average morphology across tumour types. All measurements are normalised to the data average
and standard deviation. B) Heterogeneity of cell morphology across different mesothelioma tumour types based on
standard deviation (SD) of z-scored single cell data. C-D) Morphological heterogeneity based on ground truth labels
(C) and predicted labels (D)}
\label{morph_fig}
\end{figure}

Pathologists assess cell morphology when diagnosing and scoring mesothelioma tumours. Therefore, we sought to investigate differences in nuclear morphology between mesothelioma tumours with different diagnoses. We focused on key features assessed qualitatively by pathologists including nuclear area, elongation (width and length) and nucleus shape regularity as measured by both circularity (how close it is to a circular shape), and solidity (which reflects overall concavity of a shape). Interestingly, in Figure \ref{morph_fig}, we found sarcomatoid tumours tend to have larger nuclei on average compared to epithelioid tumours. As expected these nuclei are more elongated and have less circular shapes. For almost all features, measures of nucleus shape in biphasic tumours fall in between epithelioid and sarcomatoid tumours. These results already confirm that our image analysis pipeline reflects inherent differences between these types even when only considering the average measures of each tumour which is consistent with pathological features. This motivates the development of more sophisticated AI methods to detect these differences. 

Next, we investigated the extent of variability in morphological measurements across different tumour types. We measured the standard deviation of cell features for each single tumour core as a proxy of heterogeneity. We found that Sarcomatoid tumours exhibit higher morphological heterogeneity in all nuclear features. These analyses motivate the investigation of single cell phenotypes to identify the most relevant subpopulations. 

We can gain further insight into the differences in morphology that the model is associating with each subtype by looking at the principal components of cells assigned the highest and lowest scores (i.e most and least likely to be associated with a sarcomatoid core, respectively).



%

Comparing the first principal component for each subtype in supplementary Fig. 2, we can see that the model has learnt to assign a higher score to cells with a more elongated morphology. This reflects a known distinguishing feature of the sarcomatoid morphology, validating our model scoring. This is further illustrated in the scatter plot in the upper half of Fig. \ref{morph_fig}, where we show a supervised UMAP \cite{McInnes2018} of the principal components of high and low scoring cells. The supervisory signal is provided by the output of our model, as a binary label of whether it is in the top or bottom 10\% of cells by score. Each point in the map represents a cell, colored red if it is in the top 10\% of model scores, or blue if in the bottom 10\%. UMAP attempts to learn an embedding in which examples with similar features are closer together. Thus, by looking at groups of cells in this map we can understand how high scoring and low scoring cell populations look. From Fig \ref{morph_fig}, sarcomatoid groups B and E we can see elongated cells are scored highly sarcomatoid, as are groups C and D showing large, irregular cells. Cells scored in the bottom 10\% tend to be much smaller, as can be seen in groups F, H, I and J. They also are rounder and more regular in their shape. As can be seen comparing epithelioid group G to sarcomatoid group C, while large cells may also be scored as epithelioid, they have a round shape with less texture to the staining.

\subsection{Survival Analysis using MesoScores}

\begin{figure*}[ht]

  \centering
 \centerline{\includegraphics[width=15cm]{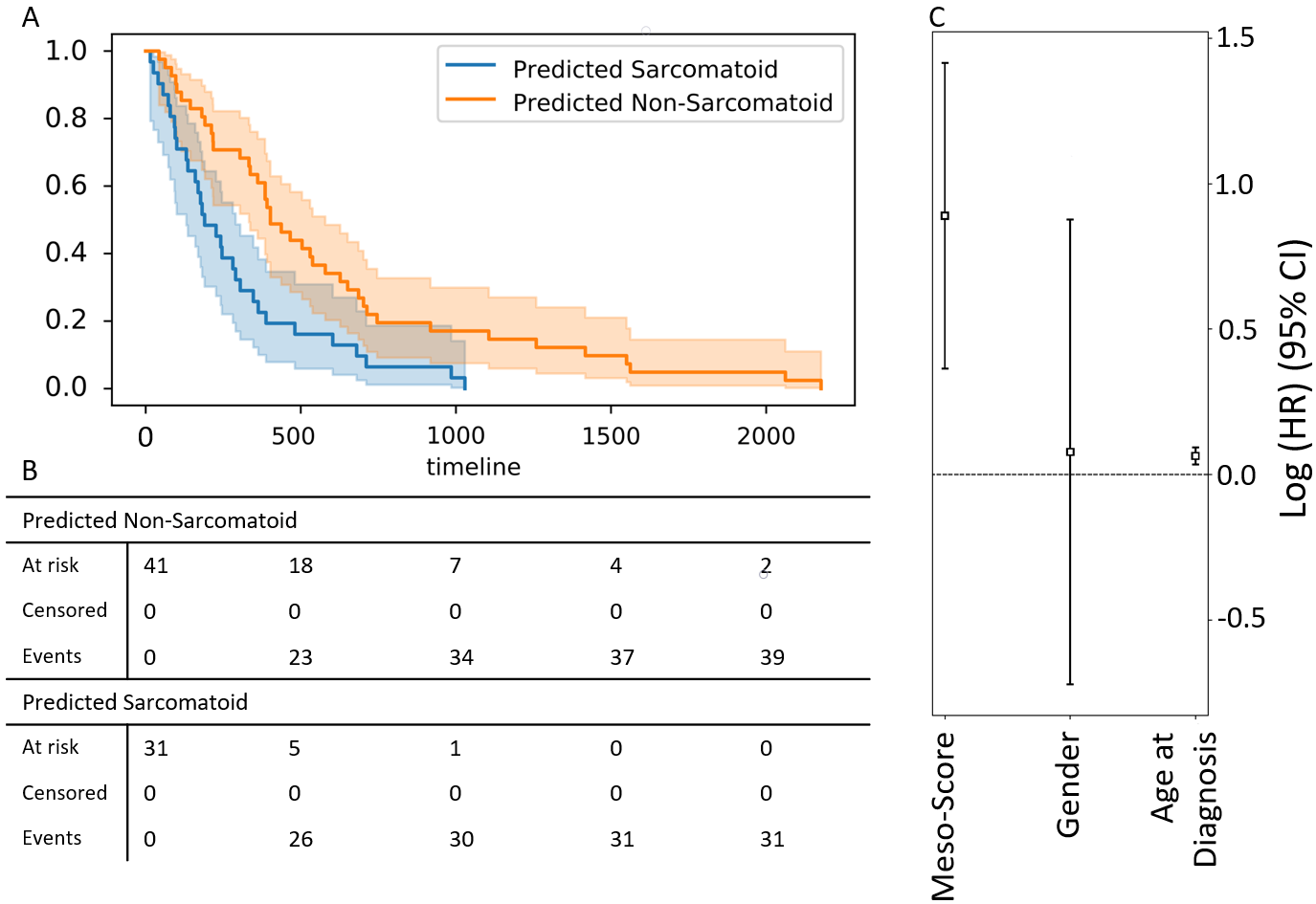}}

\caption{a) Kaplan-Meier curves for all data.  For data right censored at 3 years see supplementary material Fig 1. c) log Hazard Ratio of high Meso-score compared to demographic factors.}
\label{surv}
\end{figure*}

Survival analysis using Kaplan Meier plots are shown in Figure \ref{surv}. Patients were divided into two groups based on model score. The median survival time of patients predicted more sarcomatoid was significantly shorter compared to non-sarcomatoid patients (190 vs 402 days, $p < 0.002$) for both. This difference in survival can be observed in the Kaplan Meier plot (a), where the predicted non-sarcomatoid curve in orange is less steep than the blue predicted sarcomatoid curve. In a Cox-proportional hazard model adjusted for sex and age at diagnosis, the hazard ratio for sarcomatoid cases was 2.43 (95\% CI 1.44 – 4.12, $p<0.005$), indicating that patients with sarcomatoid morphology were 2.43 times more likely to have died at a specific time point than non-sarcomatoid cases. In comparison, the HR for both gender and age were both much smaller, at $<1.1$. Very similar findings were obtained with censoring at 3 years (see supplementary material, Fig. 1). 

\subsection{General Discussion and Future Work}

We have developed a model capable of learning a cell-level indication of sarcomatoid and epithelioid regions of a TMA core tissue sample, which enables quantitative characterisation of a core according to the relative proportions of S and E components present. In summary:

\begin{enumerate}
\item 
We have developed a Graph Neural Network (GNN) model (called MesoGraph) that can predict the mesothelioma subtype of the given patient sample with high accuracy (AUC-ROC $>$ 0.85) over independent multi-centric validation using only H\&E stained images of tumor samples.
\item
For a given sample, the proposed approach can generate a quantitative assessment (called MesoScore) of where the sample stands in terms of the epithelioid to sarcomatoid spectrum. 
\item
Model predictions can be mapped onto individual cells in a given sample to generate histograms (called MesoGrams) showing the relative densities of epithelioid and sarcomatoid components within the tumor. 
\item
MesoGraph generated scores can be used as a prognostic marker for predicting disease specific survival.
\item
We show that the weakly-supervised model is able to characterize known morphological patterns of cells associated with epithelioid and sarcomatoid mesothelioma.
\item
The code and the dataset used in this study is made publicly available for further development at https://github.com/measty/meso\textunderscore graph.
\end{enumerate}

\noindent The developed approach could help pathologists to subtype a core more accurately, consistently and efficiently, and paves the way to move beyond the three type system of characterising a tumor, towards a more fine-grained characterisation that matches the underlying continuous biological expression of mesothelial tumour cells on a spectrum between epithelioid and sarcomatoid morphology.

Most MIL based methods introduced in the literature have been patch based. One such MIL approach was introduced in \cite{dsmil2021}. Here, a dual stream approach was used where the final bag score is the mean of max instance pooling and an attention based weighted average of instances attended to by the max instance.
In another approach \cite{clinical2019} large scale datasets are used to train a MIL model for tumor detection, backpropagating only the top K instances per bag. 
The CLAM algorithm \cite{lu2020data} is a further patch-based MIL method with attention that has been applied to a variety of computational pathology tasks.
As a final example of patch-based approaches, in the IDaRS algorithm proposed in \cite{idars2021} to detect key mutations on colorectal cancer, learning occurs on patches drawn using a ranking-based sampling scheme. While these MIL approaches have been developed for patch-level instances, we develop our method treating each cell as an instance, allowing us to investigate the differences in cell morphology between subtypes. This also removes the limitation on spatial resolution of predictions imposed by a patch-based approach.

Graph Neural Networks (GNNs) have also been applied in this domain. In \cite{wang2019weakly}, a GNN is used on prostate cancer TMA cores with self-supervised and morphological features for the task of classifying examples as high or low risk according to the Gleason score. GNNs are applied to whole slide images in \cite{lu2021slidegraph} by spatially clustering cells to form agglomerate nodes from which to build a slide-level graph to predict HER2 status in breast cancer. Our approach uses a dual-task formulation with ranking loss, on a cell graph, to allow better identification of regions associated with the two components that may be present in a mesothelioma core.

The results for this study show a potential for clinical implications when applied to routine diagnosis of MM. As shown in our work, there is a gradient between epithelioid and sarcomatoid MM and the various cell populations are identifiable and can be quantified using our approach. Improving identification of mesothelioma subtypes, is an essential part of diagnosis for MM. The behavior of biphasic MM is dependant on the ratio of epithelioid and sarcomatoid cells and may also be extended to other biphasic tumours. 
 
One limitation of our method is that, while we have taken care to validate our model by looking at the features that influence its predictions, and the typical morphology of cells found in epithelioid and sarcomatoid regions, we still have some issues with the interpretability of the model outputs. Not all of the features our model learns on have an obvious histomorphological counterpart. For example, if we see from a feature importance analysis that a particular resnet feature is important, it is not clear how that translates into a histomorphological biomarker that a pathologist can look for in a tissue sample. Haralick texture features are a little better as they are constructed to capture specific, well defined properties of textures, but they are still difficult to interpret in comparison to morphological features.

Future work could involve the incorporation of cell classifications, via a segmentation method such as Hovernet \cite{graham2019hover} capable of simultaneous cell segmentation and classification.  This would provide a further informative feature that may help identify cell-type specific patterns such as an association of tumor-infiltrating lymphocytes to a specific subtype. Such features could also help move away from difficult to interpret features such as resnet features, without sacrificing performance. A more extensive evaluation considering a larger dataset and including pathologist concordance studies to identify whether pathologists using such a tool would make more consistent and more accurate subtyping could also be considered.

In conclusion, we provide a novel method for more precisely characterising epithelioid and sarcomatoid cell subtypes in a quantifiable and reproducible way. Given the importance of sarcomatoid subtypes for prognosis and deciding on treatment pathway, our method may potentially offer clinical implications for patient care. Improved subtyping of MM allows for gains in both the efficiency and reliability of assessment of mesothelioma tumour cell classification by a reporting pathologist. The method we present and future work using our approach to further define epithelioid and sarcomatoid spectrum of MM, may ultimately form a basis for improving treatment and prognosis for the patient. 

\subsection*{Acknowledgements}

\noindent This project was funded by CRUK-STFC Early Detection Innovation Award.

\noindent FM and ME also acknowledge funding support from EPSRC EP/W02909X/1.

\noindent We are grateful to The London Asbestos Support Awareness Group: https://www.lasag.org.uk/, National Mesothelioma Virtual Bank: http://www.mesotissue.org/ and MesoBank UK: http://www.mesobank.com/for their support.

\subsection*{Author Contributions and Funding}

\noindent ME: Conceptualization, Machine Learning Model Design and Implementation, Analysis, Writeup.

\noindent FM: Conceptualization, Lead for Machine Learning Model and Experiment Design, Analysis, Paper Write-up, Funding acquisition, Project Supervision and Management.

\noindent JO: Conceptualization, statistical analysis, paper writing up and revisions

\noindent JLR: Conceptualization, Clinical Lead, Paper Write-up, Funding acquisition, Project Supervision and Management.

\noindent HS: Conceptualization and study design, result analysis, paper writeup and revisions, Funding acquisition. 

\noindent STM and XG: Conceptualization, paper writeup and revisions.

\noindent EK: Conceptualization, Paper structuring and revisions.

\noindent AMF, DJ, WC, MF, SP: Data Provision.

\subsection*{Declaration of Interests}

The authors declare no competing interests.

%
%
%
\bibliographystyle{splncs04}
\small{
\bibliography{cpath}
}
%

\onecolumn

\section{Star Methods}
\subsection{Data Aquisition and Patient Characteristics}

\begin{figure*}[ht!]

  \centering
 \centerline{\includegraphics[width=15cm]{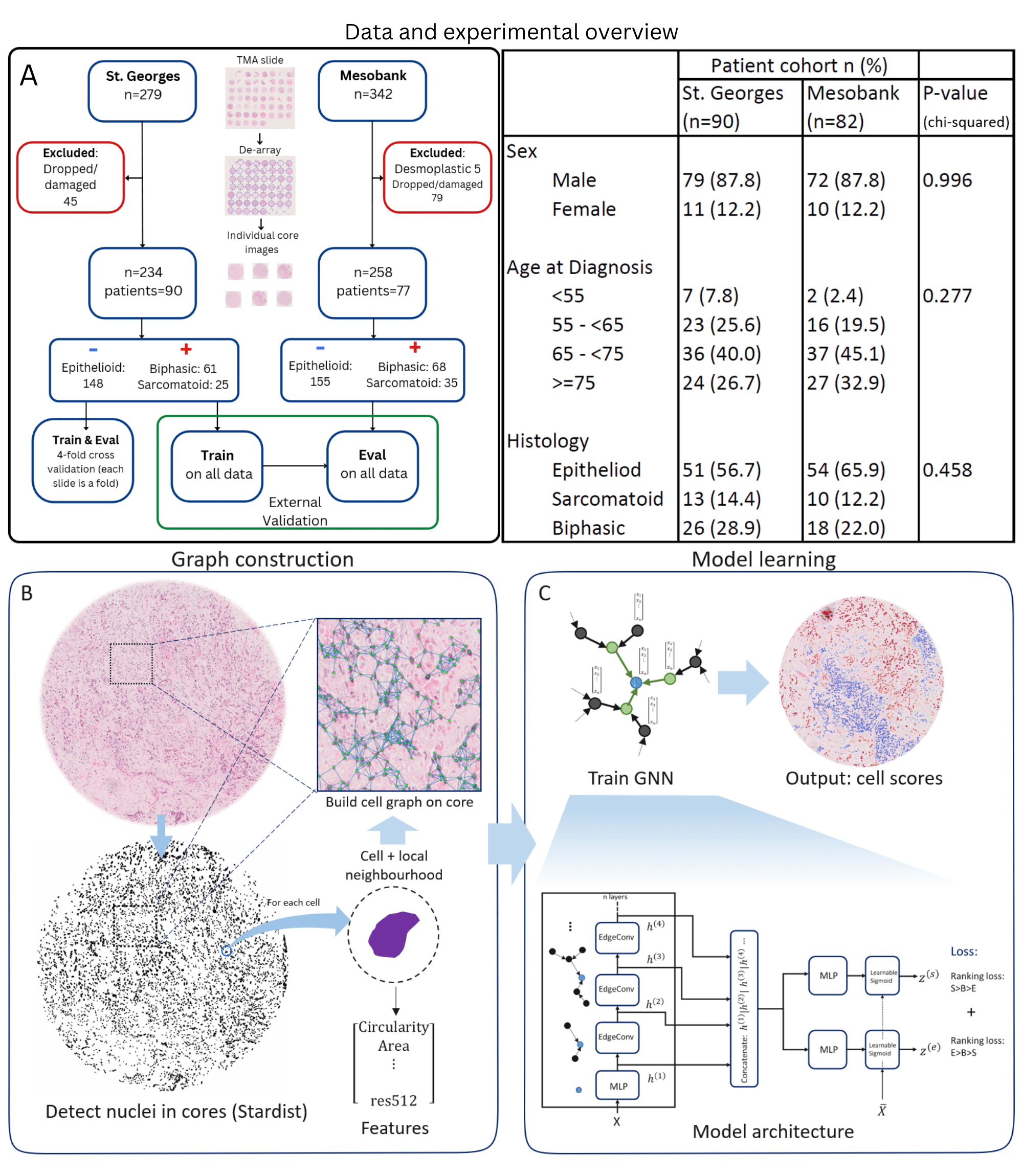}}

\caption{Study overview. A) Data \& Experimental design. TMA slides were de-arrayed into individual images, and images of cores which were dropped or particularly badly damaged were excluded. The model is trained on the St. Georges cohort and validated both internally, and on the external Mesobank cohort. 
B) Steps to represent a TMA core as a graph, from cell detection, through extraction of morphological and local neighbourhood features, to the construction of the cell graphs upon which our model will be trained. In panel C) is the Proposed MesoGraph GNN architecture. Deeper layers incorporate information from larger neighbourhoods. By concatenating layer representations, we allow the model to use information at multiple scales.}
\label{study_overview}
\end{figure*}

Two independent MM patient cohorts were obtained retrospectively (see Fig. \ref{study_overview} panel A):  The training cohort was from St. Georges Hospital and consisted of 102 patients. The validation cohort, of 82 patients, was obtained from Mesobank \cite{Rintoul380}, a UK mesothelioma biobank. Mesobank collects samples from multiple UK hospitals. The date of death for Mesobank patients had been provided by the UK National Cancer Registration and Analysis Service (NCRAS). 

The primary dataset used in this work is a collection of H\&E stained Tissue Micro-arrays (TMAs) of tumor tissue biopsies collected from St. Georges Hospital, London. It consists of 4 Tissue Micro-array (TMA) slides scanned using a Hamamatsu Nanozoomer S360 scanner at 20$\times$ (0.4415 microns per pixel) with a total of 279 cores covering 102 separate cases (patients). After the removal of dropped and severely damaged/incomplete cores, we are left with 234 cores over 90 patients, of which 148 are EM, 61 BM, and 25 SM. We additionally use a validation set of TMA cores over two slides provided by Mesobank, scanned at 20$\times$ (0.5015 microns per pixel) using a Leica Aperio AT2. The class counts after removal of dropped/damaged cores were 258 cores over 77 patients, with 155 EM, 68 BM, and 35 SM. Only core-level labels are available. We first perform Vahadane stain normalisation \cite{Vahadane2016} to minimise systematic stain variability between slides and cores. To represent a TMA core as a graph suitable for learning a GNN, we detect cells and extract features from these as described in Section \ref{GNN}.

\subsection{Problem formulation} \label{formulation}
As the biphasic subtype is a mix of epithelioid and sarcomatoid components, and subtype labels are only available at the core level, we model the subtype prediction task as a binary Multiple Instance Learning (MIL) problem. Under the MIL paradigm \cite{mil1997}, an example is represented by a bag of instances, and a bag is considered positive if it contains at least one positive sample. We express the subtyping problem as a dual MIL prediction task. In the first task, SM is considered the positive instance, whereas in the second task EM is considered the positive instance. Formulating the problem as two parallel MIL tasks allows the possibility for some instances to be negative instances in both tasks, in contrast to viewing any instance that is not sarcomatoid as being epithelioid which would be implicitly assumed in any single-task MIL formulation.

The goal of a MIL predictor is to use training data consisting of bags with bag level labels only to predict both bag and instance level labels in testing. Formally, let $\mathcal{B}=\left\{X_1,...,X_{n_\mathcal{B}}\right\}$ be a bag corresponding to a single TMA core in our dataset, where $X_i$ are instances (cells) within the bag. The number of instances $n_\mathcal{B}$ can vary across bags. Each core, represented by bag $\mathcal{B}$, is associated with a label $Y_\mathcal{B}\in\{0,1,2\}$ in the training dataset. In our formulation, considering SM as the positive instance, epithelioid-labelled cores take the label ($Y_\mathcal{B}=0$), and biphasic and sarcomatoid cores take the label ($Y_\mathcal{B}=1, Y_\mathcal{B}=2$) respectively, as we expect progressively more sarcomatoid instances in BM and SM examples. Conversely, in the dual task (where EM is considered the positive class), biphasic and epithelioid cores take the bag labels ($Y_\mathcal{B}=1, Y_\mathcal{B}=2$), with sarcomatoid becoming the negative example ($Y_\mathcal{B}=0$). This labelling system, by predominance of positive instances, is a departure from that typically used in the MIL setting, where only positive ($Y_\mathcal{B}=1$) and negative ($Y_\mathcal{B}=0$) bags exist. We deal with this with our use of a ranking-based loss, as detailed in Section \ref{model}. Our goal is then to build a machine learning model $F(\mathcal{B};\Phi)$ with trainable parameters $\Phi$ that can use a labelled training dataset $D=\{(\mathcal{B}_1,Y_1), (\mathcal{B}_2,Y_2),...,(\mathcal{B}_M,Y_M)\}$ to generate a predicted label for a test core $\mathcal{B}$. This is done by aggregating instance level predictions $z_i = g(X_i;\phi)$ to give $Z_B=F(\mathcal{B};\Phi)=Agg(\{z_i = g(X_i;\phi)|X_i\in \mathcal{B}\})$ through an appropriate aggregation function $Agg(\cdot)$ such as max or average across top most positive instances. 

Modelling the mesothelioma subtyping problem through MIL allows us to use core-level labels to learn an instance-level scoring, with which we can identify predominantly EM or SM regions in a core. This enables us to quantify where each tissue sample falls in the EM-to-SM continuum according to the relative proportions of SM and EM instances. This fine-grained and natural characterisation of a tumor can lead to more informed decisions regarding treatment.

\subsection{Building Graph Neural Networks on Tissue Cores} \label{GNN}

A tissue sample can be described by its individual component cells and their spatial arrangement within the sample. Their physical proximity will result in nearby cells affecting each other, through their shared micro-environment and interaction via various biological processes. Therefore, a natural way to represent the sample is as a graph, with each cell being a node in the graph, connected to other nearby cells in its neighbourhood.
Let $G=(V,E)$ denote a graph, where $V$ and $E$ are the sets of nodes and edges respectively. Each node $v \in V$ is associated with a feature vector $X_v$. In our case, each node $v$ is a cell, with features $X_v$ describing characteristics of the cell and its immediate surroundings. 

We use Stardist \cite{stardist2018} within QuPath \cite{qupath2017} to perform cell detection on the TMA cores. Stardist is an approach to cell detection which uses star-convex polyhedra to represent objects. For each pixel, the distances to the boundary of its containing object  along a set number of radial directions are learnt. 

For each detected cell, we use QuPath to extract features describing both the cell, and the region surrounding it, including some haralick texture features as described in \cite{haralick1973tf}, for a total of 157 features as described below.

%

\begin{itemize}
\item
Shape features: Area, length, circularity, Max and Min diameter for both nucleus and whole cell
\item
Intensity features: Mean, Median and Standard Deviation for hematoxylin and eosin channels over cell nucleus, cell cytoplasm and whole cell
\item
Shape/intensity smoothed: Above features smoothed over nearby cells using a gaussian kernel of diameter 50 $\mu$m
\item
Delaunay cluster features: number of neighbours, edge length statistics, cluster means of above features.
\item
Haralick texture features on a small circular region around detection: calculated on the eosin channel, the hematoxylin channel and on the OD sum.
\end{itemize}

In addition to these features, we extract $72\times72$ image patches centred at the centroid of each cell and use a resnet34 (imagenet pretrained weights) to extract a further 512 features for each cell, taken from the penultimate layer output of the resnet model.
We then construct the graph by connecting cells to each other cell whose centroid lies within a small radius, which we set at 30 $\mu$m. The process of building the cell graph is illustrated in Fig. \ref{study_overview}B.

\subsection{GNN Model Architecture} \label{model}

Graph neural networks (GNNs) are a powerful tool for representation learning on graphs. GNNs typically follow a neighbourhood aggregation strategy \cite{powerfulGNN2018}, where we update the representation of a node iteratively by a learned aggregation of the representations of its neighbours.
To learn the dual MIL task as described in Section \ref{formulation}, our architecture branches after the neighborhood aggregation layers. We denote the branches as Sarcomatoid (S) and Epithelioid (E) branch after the instances considered as positive in each task. An illustration of our GNN architecture can be found in panel C of Fig. \ref{study_overview}. 

Different GNN implementations vary in how they perform this aggregation, and how they combine the aggregation with the nodes current representation. We use the EdgeConv approach to aggregation from \cite{edgeconv2018}, which at layer $k>1$ takes the form:
\begin{equation}
h_v^{(k)} = \frac{1}{|N_v|}\sum_{u \in N_v}f_{\Theta_k}\left(h_v^{k-1} || h_v^{(k-1)}-h_u^{(k-1)}\right)
\end{equation}
here, $N_v$ is the neighbourhood of node $v$ (i.e the set of all nodes to which $v$ is connected), $||$ denotes concatenation, and $f_{\Theta_k}$ is chosen to be a multi-layer perceptron (MLP) with parameters $\Theta_k$. The feature representation at each layer is $h_v^{(k)} \in \mathbf{R}^{d_k}$ and the initial representation of the node is the feature vector, $h_v^{(0)} = X_v \in \mathbf{R}^{d_0}$. The output of the first layer is a purely local transformation $h_v^{(1)}=f_{\Theta_1}(X_v)$, where again $f_{\Theta_1}$ is an MLP with parameters $\Theta_1$.
At each layer we choose $f_{\Theta_k}$ to be an MLP with one hidden layer, $\text{MLP}(d_{k-1},d_k)$ with input dimension $d_{k-1}$ and hidden layer and output dimension $d_k$.
Rather than computing the final output $z_v$ at a node from the representation in the final layer only, we follow the concatenation approach in Jumping Knowledge Networks \cite{jumping2018} to combine the representation at different layers.





This combined representation from the graph convolution layers is passed to the E and S branches, to give for the S branch:

\begin{equation}
z_v^{(s)} =\sigma\left(\alpha^{(s)} f_{\Theta_s}( [h_v^{(1)} || ... || h_v^{(K)}])+\beta^{(s)} \right)
\end{equation}
Here $\sigma(\cdot)$ denotes a sigmoid function, and both $\alpha^{(s)}=f_{\alpha}^{(s)}(\bar{X})$ and $\beta^{(s)}=f_{\beta}^{(s)}(\bar{X})$ are the output of further small MLPs taking as input a core-level feature mean $\bar{X}=\frac{1}{N}\sum_{v \in V}X_v$. In a similar way, we also obtain $z_v^{(e)}$ for the E branch. 
We take the graph level prediction to be $Z=\frac{1}{|V|}\sum_{v \in V}(z_v^{(s)}-z_v^{(e)})$, the mean of the cell-level scores.

To train our model, we use a pairwise ranking loss:

\begin{equation}
L = \sum_{i \in \text{Batch}}\sum_{j \in \text{Batch}} \text{max}\left(0, 1-(Y_i-Y_j)(Z_i-Z_j)\right)
\end{equation}

where one prediction head (treating S as the positive instance) is trained to rank bags $S>B>E$ and the second is trained to rank $E>B>S$, i.e treating E as the positive instance. Our model is implemented using the PyTorch geometric framework. We used 5 EdgeConv layers, each learning a feature representation of dimension 10. We use the Adam optimiser \cite{kingma2017adam} and a decaying cyclic learning rate scheme \cite{smith15cyclic} with min and max learning rate $2\times10^-5$ and $1\times10^-4$. The cycle length is 50 epochs and at each cycle, the max lr decays by a factor of $0.8$. We train for a maximum of 500 epochs with early stopping.

\subsection{Cell Morphology Characterization}

To investigate the typical cell morphologies and morphological differences of cells assigned high and low scores by the model, our approach is similar in concept to the `eigenfaces' decomposition in \cite{turk1991eigenface}. We have taken the highest scoring 10\% of cells from sarcomatoid cores, and the lowest scoring 10\% of cells from epithelioid cores, and aligned the images of all the cells so that the major axis is oriented vertically. We have then masked out all but a small region around the cell so that as little background as possible remains.
Finally, we have taken the H channel of the aligned cell images, and performed Principal Component Analysis (PCA) on the pixel values. This process, and some of the resulting components are illustrated in supplementary Fig. 2. 

We use this analysis to illustrate the differences in morphology between cells scored highly sarcomatoid or non-sarcomatoid by our model, as presented in Section \ref{pca-umap}.

\subsection{Model Performance and Evaluation} \label{evaluation}

For performance evaluation on the primary cohort, we employ a hold-one-out cross-validation strategy over slides, so that for each fold all cores of a single slide are held out as the test set. This is done to avoid any potential bias from systematic differences between slides, and to ensure no mixing of cores from the same patient occurs between the training and testing sets. The cores to be used for training are split 75\%-25\% into train and validation sets, respectively. We compared our model with CLAM \cite{lu2020data}, and PINS \cite{eastwood2022pins}, two patch-based methods which attempt to focus training in an adaptive way on the most important instances. We additionally compared with two simple MIL approaches, max-MIL and naive-MIL. Max-MIL is a patch-based method where we backpropagate only on the maximal instance during training. This has been used in for example \cite{clinical2019}. Naive-MIL is a naive approach whereby we simply assign the bag level label to all instances in a bag, and treat all instances equally during training. For both of these methods we used a resnet34 pre-trained on imagenet as the base patch level model.
To evaluate performance on the external validation cohort, we have trained our model on the entire St. Georges cohort, and evaluated model predictions on the Mesobank cohort. Conversely, we also present results obtained training on Mesobank data and evaluate that model on the St. Georges data. Model performance is summarized in Table \ref{restable}, and Fig. \ref{model_auc}. as described in Section \ref{performance_results}.

\end{document}


\noindent Supplementary material for paper entitled `MesoGraph: Automatic Profiling of Malignant Mesothelioma Subtypes from Histological Images'

\begin{figure*}[ht]

  \centering
 \centerline{\includegraphics[width=14cm]{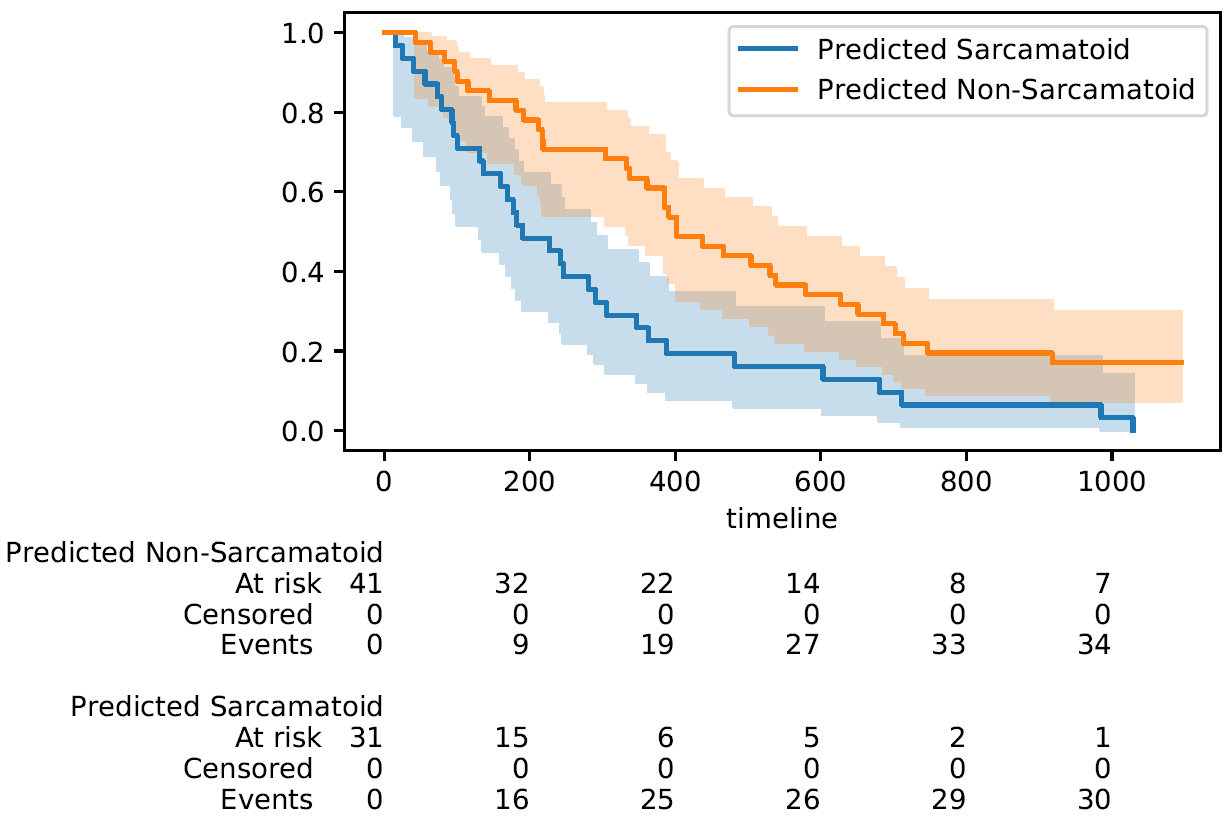}}

\caption{Kaplan-Meier curves for survival prediction when stratifying by model output score. Data right censored at 3 years.}
\label{pca}
\end{figure*}
\begin{figure*}[h]

  \centering
 \centerline{\includegraphics[width=14cm]{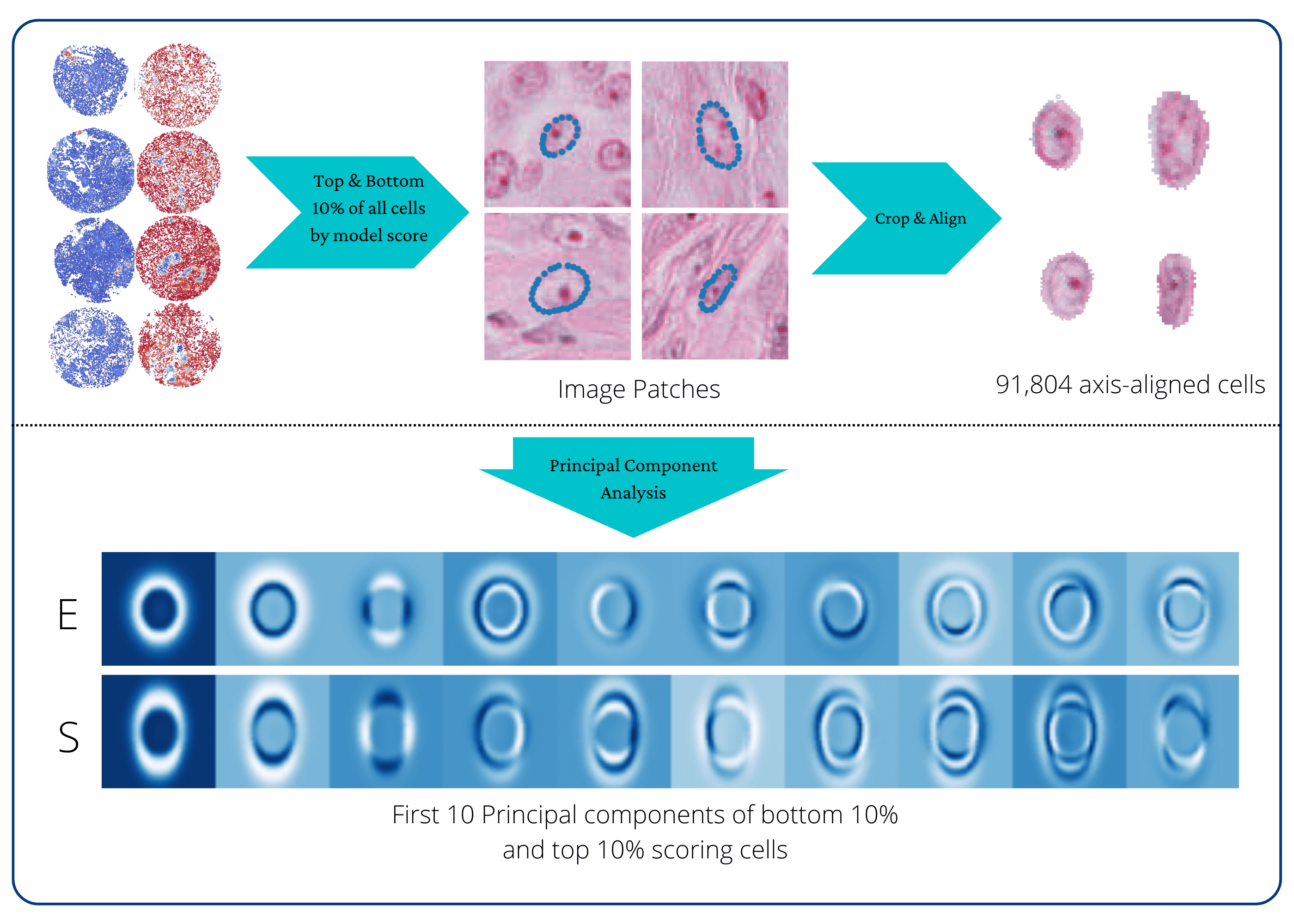}}

\caption{Principal component analysis on the top \& bottom 10\% of cells by model score. Top and bottom scoring cell populations were oriented so that their major axes are oriented in the same direction, and cropped to remove as much background as possible. The resulting PCA components show highly sarcomatoid-scoring cells align with known morphological features of sarcomatoid subtype.}
\label{pca}
\end{figure*}